\documentclass{INTERSPEECH2023}
\usepackage{latexsym}
\usepackage{csquotes}
\usepackage{amsmath}
\usepackage{amssymb}
\usepackage{graphicx}
\usepackage{subfigure}
\usepackage{wasysym}

\usepackage{algorithm}
\usepackage{algpseudocode}
\usepackage{silence}
\usepackage{pgfplots,pgfplotstable}
\tikzset{
  font={\fontsize{8pt}{10}\selectfont}}

\usepackage{multirow,ctable}

\newcommand{\sara}{\textcolor{black}}
\interspeechcameraready

\title{\sara{\textsc{AlignAtt}: Leveraging Attention-based Audio-Translation Alignments\\for Simultaneous Speech Translation}}

\title{\textsc{AlignAtt}: Using Attention-based Audio-Translation Alignments\\as a Guide for Simultaneous Speech Translation}

\name{Sara Papi$^{\Join,\diamondsuit}$, Marco Turchi$^\natural$, Matteo Negri$^\Join$}
\address{
  $^{\Join}$Fondazione Bruno Kessler, Italy\\
  $^\diamondsuit$University of Trento, Italy\\
  $^\natural$Indipendent Researcher}
\email{\{spapi,negri\}@fbk.eu, marco.turchi@gmail.com}

\begin{document}

\maketitle

\newcommand\blfootnote[1]{%
  \begingroup
  \renewcommand\thefootnote{}\footnote{#1}%
  \addtocounter{footnote}{-1}%
  \endgroup
}
\blfootnote{We acknowledge the support of the PNRR project FAIR - Future AI Research (PE00000013), under the NRRP MUR program funded by the NextGenerationEU.}
 
\begin{abstract}
Attention is the core mechanism of today's most used architectures for natural language processing and has been analyzed from many perspectives, including its effectiveness for machine translation-related tasks. Among these studies, attention resulted to be a useful source of information to get insights about word alignment also when the input text is substituted with audio segments, as in the case of the speech translation (ST) task. In this paper, we propose \textsc{AlignAtt}, a novel policy for simultaneous ST (SimulST) that exploits the attention information to generate source-target alignments that guide the model during inference. Through experiments on the 8 language pairs of MuST-C v1.0, we show that \textsc{AlignAtt} outperforms previous state-of-the-art SimulST policies applied to offline-trained models with gains in terms of BLEU of 2 points and 
% a latency reduction of 0.5-0.8$s$ among the languages.
latency reductions ranging from 0.5$s$ to 0.8$s$ across the 8 languages.
\end{abstract}

\section{Introduction}

%Recently, Speech Translation (ST) has been facing the challenge of going in simultaneous, i.e. generating partial translation while continuing to ingest input audio. Due to the absence of error propagation and thanks to reduced latency, direct models have rapidly replaced cascaded pipelines and are now commonly used \cite{iwslt_2021,anastasopoulos-etal-2022-findings} to address  Simultaneous ST (SimulST). 

Simultaneous speech translation (SimulST) involves the generation,  with minimal delay, of partial translations for an incrementally received input audio. In the quest for high-quality output and low latency, recent developments 
%have 
led to the advent of direct methods, which have been demonstrated to outperform the traditional cascaded (ASR + MT) pipelines in terms of both quality and latency \cite{anastasopoulos-etal-2022-findings}.  %iwslt_2021
%In particular, there has been a focus on offline training approaches \cite{papi-etal-2022-simultaneous,liu20s_interspeech,chen-etal-2021-direct,nguyen2021empirical}, which have been shown to significantly lower the high computational and maintenance costs associated with earlier solutions that relied on optimizing dedicated models for different latency regimes \cite{ren-etal-2020-simulspeech,ma-etal-2020-simulmt,zeng-etal-2021-realtrans}.
%
Early works on direct SimulST require the training of several models which were optimized for different latency regimes \cite{ren-etal-2020-simulspeech,ma-etal-2020-simulmt,zeng-etal-2021-realtrans}, consequently resulting in high computational and maintenance costs. 
With the aim of reducing this computational burden, the use of offline-trained direct ST models for the simultaneous inference has been recently studied \cite{papi-etal-2022-simultaneous} and is becoming popular \cite{liu20s_interspeech,chen-etal-2021-direct,nguyen2021empirical} due to 
%the competitive performance achieved
its competitive performance compared to dedicated architectures specifically developed for SimulST \cite{anastasopoulos-etal-2022-findings}.
Indeed, this approach enables an offline ST model to work in simultaneous by applying, only at inference time, a so-called \textit{decision policy}, which is in charge to determine whether to emit a partial hypothesis or wait for more audio input. 
%at inference time, which is in charge to determine whether to emit a partial hypothesis or wait for more audio input. 
As a result, no specific adaptation is required either for the SimulST task or to achieve different latency regimes.

Along this line of research, we propose \textsc{AlignAtt}, a novel policy for SimulST that exploits the audio-translation alignments obtained from the attention weights of an offline-trained model to decide whether to emit or not a partial translation. 
Our policy is based on the idea that, if the candidate token is aligned with the last frames of the input audio, the information encoded can be insufficient to safely produce that token. The audio-translation alignments are automatically generated from the attention weights, whose representativeness has been extensively studied 
%for linguistically-related 
%\mn{in various}
in linguistics-related
tasks \cite{raganato-tiedemann-2018-analysis,htut2019attention,Lamarre2022}, including word-alignment in machine translation \cite{tang-etal-2018-analysis,garg-etal-2019-jointly,chen-etal-2020-accurate}. %zenkel2019adding

%To sum up, 
All in all,
the contributions of our work are the following:
\begin{itemize}
    \item We present \textsc{AlignAtt}, a novel decision policy for SimulST that guides an offline-trained model during 
    %the simultaneous
    simultaneous inference by leveraging audio-translation alignments computed from the 
    %\mn{(cross)} 
    attention weights;
    \item We compare \textsc{AlignAtt} with popular and state-of-the-art policies that can be applied to offline-trained ST models, 
    %realizing 
    achieving
    the new state of the art on all the 8 languages of MuST-C v1.0 \cite{CATTONI2021101155}, with gains of 2 BLEU points and a latency reduction of 0.5-0.8$s$ 
    %among the languages;
    depending on the target languages;
    \item The code, the models, and the simultaneous outputs are published under Apache 2.0 Licence at: \url{https://github.com/hlt-mt/fbk-fairseq}.
\end{itemize}

\section{\textsc{AlignAtt} policy}
\label{sec:policy}

%Our proposed SimulST policy named \textsc{AlignAtt} 
\textsc{AlignAtt} is based on the source audio - target text alignment obtained through the attention scores of a Transformer-based model \cite{transformer}.
In the Transformer, encoder-decoder 
%or cross attention
(or cross) attention $A_C$ is computed by applying the standard dot-product mechanism \cite{7472621} as follows:
\begin{equation*}
    A_C(Q,K,V) = softmax \left( \frac{QK^T}{\sqrt{d_k}} \right) V
\end{equation*}
where the matrices $K$ (key) and $V$ (value) are obtained from the
%output of the encoder 
encoder output
and consequently depend on the input source $\mathbf{x}$, the matrix $Q$ (query) is obtained from the output of the previous decoder layer
%layer or
(or from the previous output tokens in
%the 
case of the first decoder layer),
and consequently depends on the prediction $\mathbf{y}$, and $d_k$ is a scaling factor. 
%as in the classical attention formulation \cite{7472618}. 
Cross attention can be hence expressed as a function of $\mathbf{x}$ and $\mathbf{y}$, obtaining 
%the matrix 
$A_C(\mathbf{x}, \mathbf{y})$.
Exploiting the cross attention $A_C(\mathbf{x}, \mathbf{y})$, the alignment vector $Align$
%can be 
is computed by considering, for each token $y_i$ of the prediction $\mathbf{y}=[y_1,...,y_m]$, the index of the most attended frame (or encoder state) $x_j$ of the source input $\mathbf{x}=[x_1,...,x_n]$:
\begin{equation*}
    Align_i = \arg \max_{j}  A_C(\mathbf{x}, y_i) 
\end{equation*}
This means that, for every predicted token $y_i$, we have a unique aligned frame $x_j$ of index $Align_i$.

\begin{figure}[tb]
    \centering
    \subfigure[The emission stops when \enquote{\textit{\textbf{Ich werde heute}}} has been generated because the token \enquote{\textit{\textbf{darüber}}} (\enquote{\textit{about}}) is aligned with
    %\mn{to} 
    an inaccessible frame (in \textcolor{red}{striped red}).]{\qquad\qquad\includegraphics[width=5.5cm]{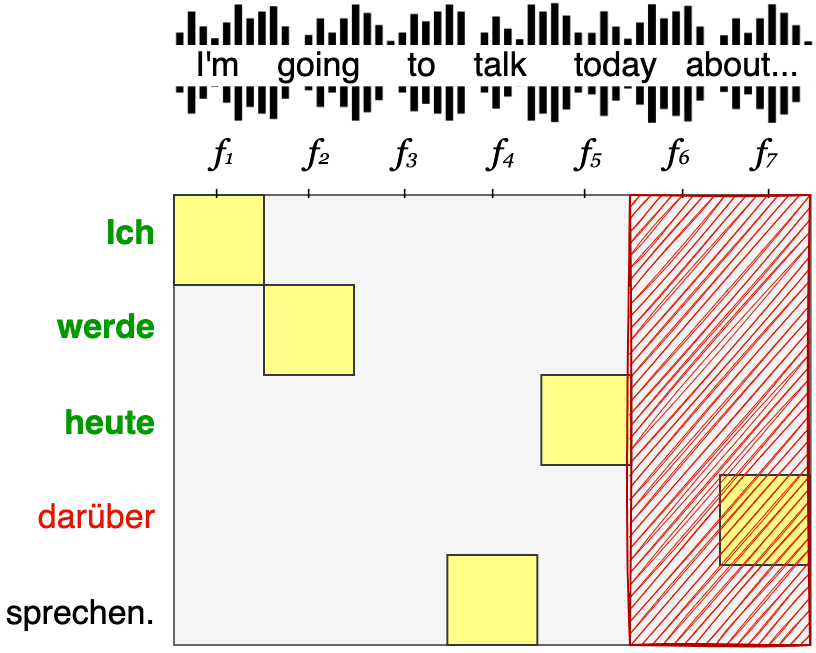}\qquad\qquad}
    \subfigure[After \enquote{\textit{Ich werde heute}}, also \enquote{\textit{\textbf{über Klima sprechen}}} is emitted since no token is aligned with
    %\mn{to}
    inaccessible frames.]{\includegraphics[width=.45\textwidth]{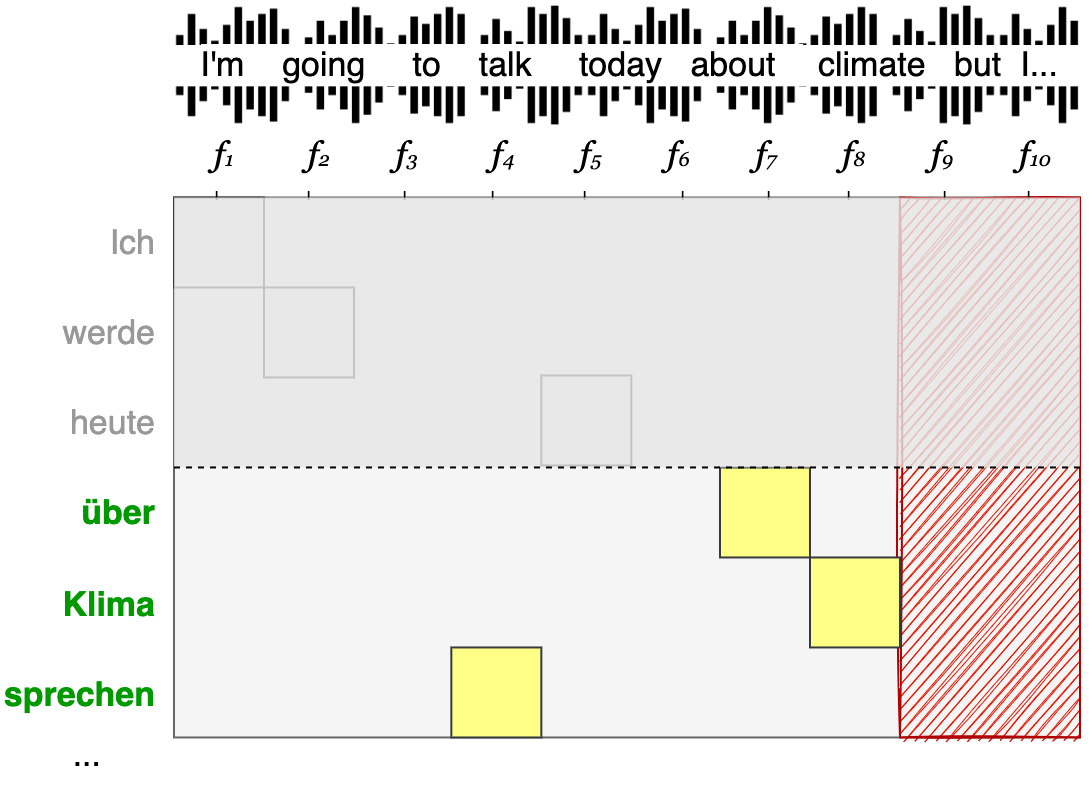}\quad}
    \caption{Example of the \textsc{AlignAtt} policy with $f=2$ at consecutive time steps $t_1$ (a) and $t_2$ (b).}
    \label{fig:alignatt}
\end{figure}

%The \textsc{AlignAtt} policy 
Our policy (Figure \ref{fig:alignatt}) 
%\mt{Non so se hai spazio, ma io aiuterei il lettore nel capire la figura 1. Io faccio onestamente fatica e se non la capisco alla prima la salto.} 
exploits the
obtained 
alignment $Align$ to guide the model during inference by checking whether each token $y_i$ attends to the last $f$ frames or 
%not and, if this condition is verified, the emission is stopped.
not. If this condition is verified, the emission is stopped, under the assumption 
%that the system will probably need more information to correctly encode the last received information contained in the last frames. 
%
%\mn{that, if attention is focused towards the most recently received speech segments, the information they provide can be insufficient to generate the hypothesis (i.e. the system has to wait for additional audio input).}
that, if a token is aligned with the most recently received audio frames, the information they provide can be insufficient to generate 
%the hypothesis
that token (i.e. the system has to wait for additional audio input).
%
%... [SPIEGHEREI LA LOGICA COME FATTO NEL PAPER DI ACL.]\sara{[Non ho capito quale parte dovrei spiegare, poco dopo il primo paragrafo dell'introduzione scrivo la logica "our policy is based on the idea that...", a cosa ti riferisci esattamente qui? Ripetere quello che ho scritto lì? Altrimenti sarebbe "con l'assunzione che gli ultimi frame possano ancora essere instabili e non rappresentare correttamente l'ultime informazioni audio ricevute" o cose così, andrebbe bene?]}. 
Specifically, starting from the first token, we iterate over the prediction $\mathbf{y}$ and continue the emission until:
\begin{equation*}
    Align_i \notin \{n-f+1, ..., n\}
\end{equation*} 
which means that we 
stop the emission as soon as we find a token that mostly attends to one of the last $f$ frames. 
%we continue the emission if the cross attention mostly attends to the frames before the last $f$ frames. 
Thus, $f$ is the parameter that directly controls the latency of the model: smaller $f$ values mean fewer frames to be considered inaccessible by the model, consequently implying 
%lower latency and vice versa.
a lower chance that our stopping condition is verified and, in turn, lower latency.
The process is formalized in Algorithm \ref{alg:alignatt}.
\begin{algorithm}
\caption{\textsc{AlignAtt}}\label{alg:alignatt}
\begin{algorithmic}
\Require $Align$, $f$, $\mathbf{y}$ 
\State $i \gets 1$
\State $prediction \gets [\quad]$
\State $stop \gets False$
\While{$stop \neq True$}
\If{$Align_i \in \{n-f+1,...,n\}$}
    \State $stop \gets True$  \Comment{inaccessible frame}
\Else
    \State $prediction \gets prediction + y_i$
    \State $i \gets i + 1$
\EndIf
\EndWhile
\end{algorithmic}
\end{algorithm}

Since in SimulST the source speech input $\mathbf{x}$ is incrementally 
 received and its length $n$ is increased at every time step $t$, applying the \textsc{AlignAtt} policy means applying Algorithm \ref{alg:alignatt} at each timestep to emit 
 (or not) 
 %\mn{(or retain)}
 the partial hypothesis until the input $\mathbf{x}(t)$ has been entirely received. 

\section{Experimental Settings}

\subsection{Data}
\label{subsec:data}
We train one model for each of the 8 languages of MuST-C v1.0~\cite{CATTONI2021101155}, namely English (en) to Dutch (nl), French (fr), German (de), Italian (it), Portuguese (pt), Romanian (ro), Russian (ru), and Spanish (es). 
We
%\mn{During training, we}
filter out segments longer than 30$s$ from the training set to optimize GPU RAM consumption.
%, the resulting data statistics are presented in Table \ref{tab:data}.
%\mt{``optimize GPU RAM consumption'' non vorrei che attirasse la domanda del reviewer, del perche' e per come questo accade. Mi domando se va specificato il perche' lo fai.}, 
%
%\begin{table}[htb]
%    \centering
%    \setlength{\tabcolsep}{4pt}
%    \begin{tabular}{cccccccc}
%    \specialrule{.1em}{.05em}{.05em} 
%        de & es & fr & it & nl & pt & ro & ru \\
%        \hline
%        225K & 260K & 269K & 248K & 244K & 201K & 231K & 260K \\
%    \specialrule{.1em}{.05em}{.05em} 
%    \end{tabular}
%    \caption{Number of sentences of the training set for each language of MuST-C v1.0.}
%    \label{tab:data}
%\end{table}
%
We also apply sequence-level knowledge distillation \cite{kim2016sequencelevel} to 
%enhance the performance. We employ
increase the size of our training set and improve performance. To this aim, we employ
NLLB 3.3B \cite{costa2022no} as the MT model to translate the English transcripts of the 
%train
training set into each of the 8 languages, and we use the automatic translations together with the gold ones during 
%training, therefore
training. As a result,
%. This means that 
the final number of target sentences 
%used for training 
is 
%double 
twice
%that shown in Table \ref{tab:data}, 
the original one %,
while the speech input remains unaltered.
The performance of the NLLB 3.3B model on the MuST-C v1.0 test set is shown in Table \ref{tab:nllb}. 

\begin{table}[!htb]
%\small
    \centering
    \setlength{\tabcolsep}{3pt}
    \begin{tabular}{lccccccccc}
    \specialrule{.1em}{.05em}{.05em} 
        Model & de & es & fr & it & nl & pt & ro & ru & Avg \\
        \hline
        NLLB & 33.1 & 38.5 & 46.5 & 34.4 & 37.7 & 40.4 & 32.8 & 23.5 & 35.9 \\
    \specialrule{.1em}{.05em}{.05em} 
    \end{tabular}
    \caption{BLEU results on all the language pairs of MuST-C v1.0 tst-COMMON of NLLB 3.3B model.
    %Results of NLLB 3.3B on MuST-C v1.0.
    }
    \label{tab:nllb}
\end{table}

\subsection{Architecture and Training Setup}
\label{sec:architecture}
The model is made of 12 Conformer \cite{gulati20_interspeech} encoder layers and 6 Transformer decoder layers, 
%each one 
having 8 attention heads each. The embedding size is set to 512 and the feed-forward layers are composed of 2,048 neurons, with $\sim$115M parameters in total.
The input is represented by 80 log Mel-filterbank audio features extracted every 10$ms$ with a sample window of 25, and pre-processed by two 1D convolutional layers of striding 2 to reduce the input length by a factor of 4 \cite{wang2020fairseqs2t}. 
%\mt{Usi sia le convoluzioni e la CTC per ridurre l'input. Credevo che le due cose fossero exclusive, ovvero la CTC permetteva di non dover ridurre l'input all'inizio del dell'encoer. Se e' una cavolta, ignorala pure.} 
Dropout is set to 0.1 for attention, feed-forward, and convolutional layers. The kernel size is 31 for both point- and depth-wise convolutions in the Conformer encoder. The SentencePiece-based \cite{sennrich-etal-2016-neural} vocabulary size is 8,000 for translation and 5,000 for transcript. Adam optimizer with label-smoothed cross-entropy loss (smoothing factor 0.1) is used during training together with CTC loss \cite{Graves2006ConnectionistTC} to compress audio input representation and speed-up inference time~\cite{gaido-etal-2021-ctc}. Learning rate is set to $5\cdot10^{-3}$ with Noam scheduler and 25,000 warm-up steps.
Utterance-level Cepstral Mean and Variance Normalization (CMVN) and SpecAugment \cite{Park2019} are also applied during training.  
Trainings are performed on 2 NVIDIA A40 GPUs with 40GB RAM. We set 40k as the maximum number of tokens per mini-batch, update frequency 4, and 100,000 maximum updates ($\sim$28 hours). 
%We also earlier stop training if validation loss does not improve for 10 epochs. 
Early stopping is applied during training if validation loss does not improve for 10 epochs.
We use the bug-free implementation of fairseq-ST \cite{papi2023reproducibility}.

\begin{table*}[!htb]
%\small
    \centering
    \setlength{\tabcolsep}{8pt}
    \begin{tabular}{lccccccccccc}
    \specialrule{.1em}{.05em}{.05em} 
        \multirow{2}{*}{Model} & \multicolumn{2}{c}{Ext. Data} & \multirow{2}{*}{de} & \multirow{2}{*}{es} & \multirow{2}{*}{fr} & \multirow{2}{*}{it} & \multirow{2}{*}{nl} & \multirow{2}{*}{pt} & \multirow{2}{*}{ro} & \multirow{2}{*}{ru} & \multirow{2}{*}{Avg} \\
        \cline{2-3}
         & Speech & Text & & & & & & & \\
        \hline
        %Adapted Transformer \cite{gangi19_interspeech} & $\checkmark$ & - & 17.3 & 20.8 & 26.9 & 16.8 & 18.8 & 20.1 & 16.5 & 10.5 \\
        Fairseq-ST \cite{wang2020fairseqs2t} & - & - & 22.7 & 27.2 & 32.9 & 22.7 & 27.3 & 28.1 & 21.9 & 15.3 & 24.8 \\ 
        ESPnet-ST \cite{inaguma-etal-2020-espnet} & - & - & 22.9 & 28.0 & 32.8 & 23.8 & 27.4 & 28.0 & 21.9 & 15.8 & 25.1 \\
        %NeurST \cite{zhao-etal-2021-neurst} & $\checkmark$ & - & 22.8 & 27.4 & 33.3 & 22.9 & 27.2 & 28.7 & 22.2 & 15.1 & 25.0 \\
        Chimera \cite{han-etal-2021-learning} & $\checkmark$ & $\checkmark$ & 27.1 & 30.6 & 35.6 & 25.0 & 29.2 & 30.2 & 24.0 & 17.4 & 27.4 \\
        W-Transf. \cite{ye21_interspeech} & $\checkmark$ & - & 23.6 & 28.4 & 34.6 & 24.0 & 29.0 & 29.6 & 22.4 & 14.4 & 25.8 \\
        XSTNet \cite{ye21_interspeech} & $\checkmark$ & $\checkmark$ & 27.8 & 30.8 & 38.0 & 26.4 & 31.2 & 32.4 & 25.7 & \textbf{18.5} & 28.9 \\
        LNA-E,D \cite{li-etal-2021-multilingual} & $\checkmark$ & $\checkmark$ & 24.3 & 28.4 & 34.6 & 24.4 & 28.3 & 30.5 & 23.3 & 15.9 & 26.2 \\
        LightweightAdaptor \cite{le-etal-2021-lightweight} & - & - & 24.6 & 28.7 & 34.8 & 25.0 & 28.8 & 31.0 & 23.7 & 16.4 & 26.6 \\
        E2E-ST-TDA \cite{Du_Zhang_Wang_Chen_Xie_Xu_2022} & $\checkmark$ & $\checkmark$ & 25.4 & 29.6 & 36.1 & 25.1 & 29.6 & 31.1 & 23.9 & 16.4 & 27.2 \\ 
        STEMM \cite{fang-etal-2022-stemm} & $\checkmark$ & $\checkmark$ & \textbf{28.7} & 31.0 & 37.4 & 25.8 & 30.5 & 31.7 & 24.5 & 17.8 & 28.4 \\
        ConST \cite{ye-etal-2022-cross} & $\checkmark$ & - & 25.7 & 30.4 & 36.8 & 26.3 & 30.6 & 32.0 & 24.8 & 17.3 & 28.0 \\
        \hline
        %ours (no KD) & - & - & 25.0 & 30.5 & 36.4 & 26.3 & 30.3 & 30.1 & 24.7 & 17.4 & 27.6 \\
        ours & - & $\checkmark$ & 28.0 & \textbf{31.5} & \textbf{39.0} & \textbf{27.3} & \textbf{31.8} & \textbf{32.9} & \textbf{26.3} & 18.4 & \textbf{29.4} \\
    \specialrule{.1em}{.05em}{.05em} 
    \end{tabular}
    \caption{BLEU results on MuST-C v1.0 tst-COMMON. \enquote{Ext. Data} means that external data has been used for training: \enquote{Speech} means that either unlabelled or labelled additional speech data is used to train or initialize the model, \enquote{Text} means that either machine-translated or monolingual texts are used to train or initialize the model. \enquote{Avg} means the average over the 8 languages.}
    \label{tab:offline_res}
\end{table*}

\subsection{Terms of Comparison}
\label{subsec:comparison}

We conduct experimental comparisons with the other SimulST policies that can be applied to offline systems, thus policies that do not require training 
%or
nor
adaptation to be run, namely:

\begin{itemize}
    \item \textbf{Local Agreement (LA)} \cite{liu20s_interspeech}: the policy used by \cite{polak-etal-2022-cuni} 
    % for the IWSLT 2022 Simultaneous Speech Translation Evaluation Campaign \cite{anastasopoulos-etal-2022-findings}, 
    to win 
    % \sara{the SimulST Evaluation Campaign}
    % %task 
    % at 
    % %the 
    % IWSLT 2022
    the SimulST task at the IWSLT 2022 evaluation campaign
    \cite{anastasopoulos-etal-2022-findings}.
   With this policy, a partial hypothesis is generated 
   %every
   each
   time a new speech segment is added as 
   %input and is
   input, and it is
   emitted, entirely or 
   %a part of it, 
   partially,
   if the previously generated hypothesis is equal to the current one.
    We adapted the docker released by the authors to Fairseq-ST \cite{wang2020fairseqs2t}. Different latency regimes are obtained by varying the speech segment length $T_s$.
    \item \textbf{Wait-k} \cite{ma-etal-2019-stacl}: the most popular policy originally published for simultaneous machine translation and then adapted to SimulST \cite{ren-etal-2020-simulspeech,zeng-etal-2021-realtrans}. It consists in waiting for a predefined number of words ($k$) before starting to alternate between writing a word and waiting for new output.
    We employ adaptive word detection guided by the CTC prediction to detect the number of words in the speech as in \cite{zeng-etal-2021-realtrans,papi-etal-2022-simultaneous}.
    \item \textbf{\textsc{EDAtt}} \cite{papi2022attention}: the only existing policy that exploits the attention mechanism to guide the inference. Contrary to our policy that computes audio-text alignments starting from the attention scores, in 
    %the 
    \textsc{EDAtt}
    %policy the attention scores are selected through a hyper-parameter $\lambda$ that is empirically determined on the validation set. Then, the selected attention scores are summed and a threshold $\alpha$ is used to trigger the emission. Following the authors' finding, $\lambda$ is set to 2. 
    the attention scores of the last $\lambda$ frames are summed and a threshold $\alpha$ is used to trigger the emission. While $\alpha$ handles the latency, $\lambda$ is a hyper-parameter that has to be empirically determined on the validation set. This represents the main flaw of this policy since, in theory, $\lambda$ has to be estimated for each language. Here, we set $\lambda=2$ following the authors' finding.
    %on en-de, and en-es.
    
    %\mn{[DUE PAROLE SUGLI SVANTAGGI DI EDATT, TANTO PER GIUSTIFICARE LA TUA VARIANTE E PREPARARE IL LETTORE A CONCLUSIONI/COMMENTI CHE VERRANNO NEL RESTO DEL PAPER (ALMENO CREDO)?]}
\end{itemize}

\subsection{Inference and Evaluation}
For inference, the input features are computed on the fly and Global CMVN normalization is applied as in \cite{ma-etal-2020-simulmt}. 
We use the SimulEval tool \cite{ma-etal-2020-simuleval} to 
%evaluate the policies. 
compare \textsc{AlignAtt} with the above policies.
For the LA policy, we set $T_s=[10,15,20,25,30]$\footnote{Smaller 
%$T_s$ does 
values of $T_s$ do
not improve computational aware latency.}; for the wait-k, we vary $k$ in $[2,3,4,5,6,7]$\footnote{We do not report results obtained with $k=1$ since the translation quality highly degrades.}; for \textsc{EDAtt}, we set $\alpha=[0.6,0.4,0.2,0.1,0.05,0.03]$\footnote{{These are the same values indicated by the authors of the policy.}}; for \textsc{AlignAtt}, we vary $f$ in $[2,4,6,8,10,12,14]$. 
%\mt{Metterei che la miglior configurazione e' selezionata sul dev set.}
% Moreover, for our policy, we extract the attention weights from the  4\textsuperscript{th} 
% %\mn{[PERCHE' IL QUARTO? NON ANDREBBE MOTIVATO?]} 
% decoder layer and average across all the attention heads to be comparable with \textsc{EDAtt}. 
Moreover, to be comparable with \textsc{EDAtt}, for our policy we extract the attention weights from the  4\textsuperscript{th} decoder layer and average across all the attention heads.
All inferences are performed on a single NVIDIA TESLA K80 GPU with 12GB of RAM as in the IWSLT Simultaneous evaluation campaigns \cite{iwslt_2021,anastasopoulos-etal-2022-findings}.
% To evaluate translation quality, sacreBLEU \cite{post-2018-call}\footnote{BLEU+case.mixed+smooth.exp+tok.13a+version.1.5.1} is adopted while, for evaluating latency, we use Length Adaptive Average Lagging \cite{papi-etal-2022-generation} -- or LAAL, a speech version of  Average Lagging \cite{ma-etal-2019-stacl} accounting for both longer and shorter predictions compared to the reference.
We use sacreBLEU ($\uparrow$) \cite{post-2018-call}\footnote{BLEU+case.mixed+smooth.exp+tok.13a+version.1.5.1} to evaluate translation quality and  Length Adaptive Average Lagging \cite{papi-etal-2022-generation} -- or LAAL ($\downarrow$) -- to measure latency.\footnote{Length Adaptive Average Lagging is a an improved speech version of Average Lagging \cite{ma-etal-2019-stacl}, which accounts for both longer and shorter predictions compared to the reference.}
%as in the default SimulEval evaluation setup. 
As suggested by \cite{ma-etal-2020-simulmt}, we report the computational-aware version of LAAL\footnote{We present all the results with $\text{LAAL\textsubscript{max}}=3.5s$.
%We set $\text{LAAL\textsubscript{max}}=3.5s$ for the plots.
} that accounts for the real elapsed time instead of the ideal one, consequently providing a more realistic latency measure. 

\pgfplotstableread[row sep=\\]{
BLEU	AL \\
17.2	2.049 \\
21.0	2.427 \\
23.0	2.780 \\
24.4	3.057 \\
25.2    3.343 \\
}\DEwaitk

\pgfplotstableread[row sep=\\]{
BLEU	AL \\
20.3	2.631 \\
24.0	2.825 \\
25.8	3.057 \\
26.9	3.254 \\
27.1    3.494 \\
}\DEcommon

\pgfplotstableread[row sep=\\]{
BLEU	AL \\
%18.4	1.782 \\
20.3	1.920 \\
23.1	2.254 \\
25.6    2.765 \\
26.8	3.367 \\
%27.4    3.964 \\
}\DEedatt

\pgfplotstableread[row sep=\\]{
BLEU	AL \\
19.3	1.918 \\
22.7	2.077 \\
24.9	2.338 \\
26.1	2.558 \\
26.7    2.892 \\
26.9    3.157 \\
27.3    3.406 \\
}\DEalignatt

\pgfplotstableread[row sep=\\]{
BLEU	AL \\
21.6	2.098 \\
25.7	2.514 \\
27.7	2.829 \\
29.1	3.135 \\
29.7    3.411 \\
}\ESwaitk

\pgfplotstableread[row sep=\\]{
BLEU	AL \\
22.4	2.579 \\
26.8	2.772 \\
28.6	2.990 \\
29.6	3.122 \\
29.9    3.435 \\
}\EScommon

\pgfplotstableread[row sep=\\]{
BLEU	AL \\
21.4	1.885 \\
22.9	1.973 \\
24.7	2.144 \\
26.5	2.342 \\
27.9	2.626 \\
28.9    2.975 \\
}\ESedatt

\pgfplotstableread[row sep=\\]{
BLEU	AL \\
21.8	1.868 \\
25.0	2.089 \\
27.6	2.362 \\
28.8	2.650 \\
29.6	2.918 \\
30.2    3.182 \\
30.3    3.428 \\
}\ESalignatt

\pgfplotstableread[row sep=\\]{
BLEU	AL \\
28.3	2.287 \\
31.9    2.648 \\
34.2	2.949 \\
35.6    3.233 \\
36.4	3.424 \\
}\FRwaitk

\pgfplotstableread[row sep=\\]{
BLEU	AL \\
29.0    2.560 \\
34.4	2.771 \\
36.7	3.007 \\
37.6	3.135 \\
38.1	3.460 \\
}\FRcommon

\pgfplotstableread[row sep=\\]{
BLEU	AL \\
26.8	1.884 \\
29.3	1.968 \\
32.5	2.193 \\
35.5	2.512 \\
37.0	2.771 \\
37.6    3.213 \\
}\FRedatt

\pgfplotstableread[row sep=\\]{
BLEU	AL \\
28.0	1.849 \\
32.1	2.055 \\
35.2	2.332 \\
36.6	2.653 \\
37.4	2.942 \\
37.7    3.214 \\
38.0    3.443 \\
}\FRalignatt

\pgfplotstableread[row sep=\\]{
BLEU	AL \\
19.7	2.401 \\
22.4	2.771 \\
24.3    3.089 \\
25.1    3.383 \\
}\ITwaitk

\pgfplotstableread[row sep=\\]{
BLEU	AL \\
19.4	2.692 \\
22.9	2.774 \\
24.8	3.120 \\
25.8	3.301 \\
}\ITcommon

\pgfplotstableread[row sep=\\]{
BLEU	AL \\
18.2	1.970 \\
19.4	2.007 \\
21.5	2.209 \\
23.8	2.495 \\
25.1	2.720 \\
25.6    3.084 \\
}\ITedatt

\pgfplotstableread[row sep=\\]{
BLEU	AL \\
19.3	1.893 \\
22.1	2.141 \\
24.1	2.414 \\
25.2	2.698 \\
25.9	2.959 \\
26.2    3.227 \\
26.4    3.477 \\
}\ITalignatt

\pgfplotstableread[row sep=\\]{
BLEU	AL \\
22.0	2.071 \\
25.4	2.455 \\
27.5	2.695 \\
29.1    3.096 \\
29.6    3.382 \\
}\NLwaitk

\pgfplotstableread[row sep=\\]{
BLEU	AL \\
21.7	2.464 \\
26.6	2.662 \\
28.9	2.941 \\
29.6	3.172 \\
30.2    3.392 \\
}\NLcommon

\pgfplotstableread[row sep=\\]{
BLEU	AL \\
20.3	1.816 \\
23.8	1.879 \\
26.8	2.216 \\
29.0	2.755 \\
}\NLedatt

\pgfplotstableread[row sep=\\]{
BLEU	AL \\
22.6	1.871 \\
26.3	2.061 \\
28.1	2.323 \\
29.2	2.606 \\
29.8	2.835 \\
30.4    3.096 \\
30.6    3.434 \\
}\NLalignatt

\pgfplotstableread[row sep=\\]{
BLEU	AL \\
22.7	2.454 \\
25.7    2.834 \\
27.7	3.159 \\
28.9	3.462 \\
}\PTwaitk

\pgfplotstableread[row sep=\\]{
BLEU	AL \\
23.6	2.679 \\
28.0	2.868 \\
30.1	3.092 \\
31.1	3.258 \\
31.6    3.480 \\
}\PTcommon

\pgfplotstableread[row sep=\\]{
BLEU	AL \\
23.1	1.949 \\
25.0	2.084 \\
26.9	2.275 \\
29.1	2.541 \\
30.6	2.921 \\
31.4    3.306 \\
}\PTedatt

\pgfplotstableread[row sep=\\]{
BLEU	AL \\
23.4	1.969 \\
26.8	2.157 \\
29.2	2.446 \\
30.6	2.742 \\
31.0	3.027 \\
31.6    3.274 \\
}\PTalignatt

\pgfplotstableread[row sep=\\]{
BLEU	AL \\
17.7	2.085 \\
21.2    2.511 \\
23.6	2.874 \\
24.6	3.199 \\
25.2    3.372 \\
}\ROwaitk

\pgfplotstableread[row sep=\\]{
BLEU	AL \\
19.8	2.545 \\
23.3	2.841 \\
24.8	3.078 \\
25.6	3.303 \\
26.0    3.504 \\
}\ROcommon

\pgfplotstableread[row sep=\\]{
BLEU	AL \\
18.2	1.976 \\
19.7	2.030 \\
22.2	2.199 \\
23.8	2.474 \\
24.8	2.715 \\
25.4    3.137 \\
}\ROedatt

\pgfplotstableread[row sep=\\]{
BLEU	AL \\
20.3	2.068 \\
22.8	2.158 \\
24.5	2.480 \\
25.3	2.781 \\
25.8	3.030 \\
26.0    3.293 \\
26.1    3.443 \\
}\ROalignatt

\pgfplotstableread[row sep=\\]{
BLEU	AL \\
11.9    2.172 \\
13.7    2.595 \\
15.0    2.954 \\
16.5    3.293 \\
}\RUwaitk

\pgfplotstableread[row sep=\\]{
BLEU	AL \\
12.7    2.722 \\
15.2    2.922 \\
16.6    3.179 \\
17.0    3.449 \\
}\RUcommon

\pgfplotstableread[row sep=\\]{
BLEU	AL \\
12.1    1.953 \\
14.1    2.274 \\
16.0    3.014 \\
}\RUedatt

\pgfplotstableread[row sep=\\]{
BLEU	AL \\
12.8	2.049 \\
14.2	2.135 \\
15.4    2.370 \\
16.3    2.666 \\
17.0    2.986 \\
17.2    3.274 \\
}\RUalignatt

\begin{figure*}[!t]
\centering
\small
\subfigure[en$\rightarrow$de]{
\begin{tikzpicture}
    \begin{axis}[
            ymajorgrids=true,
            xtick pos=left,
            ytick pos=left,
            minor y tick num=1,
            minor x tick num=1,
            ytick={18,20,22,24,26,28},
            %extra y ticks={22, 24},
            %ymax=27,
            ymin=17,
            ymax=28,
            xmin=1.75,
            xmax=3.5,
            ylabel=BLEU, xlabel=LAAL (s),
            ylabel shift={-4pt},
            width=4.6cm,
            height=5.2cm,
            xtick=data,
            compat=newest,
            xtick={1.5,2,2.5,3,3.5,4,4.5},
            legend style={at={(0.5,-0.2)},    
                    anchor=north,legend columns=4},   
            legend to name={mylegend},
        ]
        \addplot[color=orange, mark=pentagon*] table[x=AL,y=BLEU]{\DEwaitk};
        \addplot[color=blue, mark=triangle*] table[x=AL,y=BLEU]{\DEcommon};
        \addplot[color=teal, mark=diamond*] table[x=AL,y=BLEU]{\DEedatt};
        \addplot[color=red, mark=*] table[x=AL,y=BLEU]{\DEalignatt};
    \end{axis}
\end{tikzpicture}
}
\subfigure[en$\rightarrow$es]{
\begin{tikzpicture}
    \footnotesize
    \begin{axis}[
            ymajorgrids=true,
            xtick pos=left,
            ytick pos=left,
            minor y tick num=1,
            minor x tick num=1,
            ytick={19,21,23,25,27,29,31},
            %extra y ticks={22, 24},
            ymax=31,
            ymin=21,
            xmin=1.75,
            xmax=3.5,
            ylabel=, xlabel=LAAL (s),
            width=4.6cm,
            height=5.2cm,
            compat=newest,
            xtick=data,
            xtick={1.5,2,2.5,3,3.5,4,4.5},
            legend style={at={(0.5,-0.2)},    
                    anchor=north,legend columns=4},   
            legend to name={mylegend},
        ]
        \addplot[color=orange, mark=pentagon*] table[x=AL,y=BLEU]{\ESwaitk};
        \addplot[color=blue, mark=triangle*] table[x=AL,y=BLEU]{\EScommon};
        \addplot[color=teal, mark=diamond*] table[x=AL,y=BLEU]{\ESedatt};
        \addplot[color=red, mark=*] table[x=AL,y=BLEU]{\ESalignatt};
    \end{axis}
\end{tikzpicture} 
}
\subfigure[en$\rightarrow$fr]{
\begin{tikzpicture}
    \footnotesize
    \begin{axis}[
            ymajorgrids=true,
            xtick pos=left,
            ytick pos=left,
            minor y tick num=1,
            minor x tick num=1,
            ytick={27,29,31,33,35,37,39},
            ymax=39,
            ymin=26,
            xmin=1.75,
            xmax=3.5,
            ylabel=, xlabel=LAAL (s),
            width=4.6cm,
            height=5.2cm,
            compat=newest,
            xtick=data,
            xtick={1.5,2,2.5,3,3.5},
            legend style={at={(0.5,-0.2)},    
                    anchor=north,legend columns=4},   
            legend to name={mylegend},
        ]
        \addplot[color=orange, mark=pentagon*] table[x=AL,y=BLEU]{\FRwaitk};
        \addplot[color=blue, mark=triangle*] table[x=AL,y=BLEU]{\FRcommon};
        \addplot[color=teal, mark=diamond*] table[x=AL,y=BLEU]{\FRedatt};
        \addplot[color=red, mark=*] table[x=AL,y=BLEU]{\FRalignatt};
    \end{axis}
\end{tikzpicture} 
}
\subfigure[en$\rightarrow$it]{
\begin{tikzpicture}
    \footnotesize
    \begin{axis}[
            ymajorgrids=true,
            xtick pos=left,
            ytick pos=left,
            minor y tick num=1,
            minor x tick num=1,
            ytick={17,19,21,23,25,27},
            %extra y ticks={22, 24},
            ymax=27,
            ymin=18,
            xmin=1.75,
            xmax=3.5,
            ylabel=, xlabel=LAAL (s),
            width=4.6cm,
            height=5.2cm,
            compat=newest,
            xtick=data,
            xtick={1.5,2,2.5,3,3.5},
            legend style={at={(0.5,-0.2)},    
                    anchor=north,legend columns=4},   
            legend to name={mylegend},
        ]
        \addplot[color=orange, mark=pentagon*] table[x=AL,y=BLEU]{\ITwaitk};
        \addplot[color=blue, mark=triangle*] table[x=AL,y=BLEU]{\ITcommon};
        \addplot[color=teal, mark=diamond*] table[x=AL,y=BLEU]{\ITedatt};
        \addplot[color=red, mark=*] table[x=AL,y=BLEU]{\ITalignatt};
    \end{axis}
\end{tikzpicture} 
}
\subfigure[en$\rightarrow$nl]{
\begin{tikzpicture}
    \footnotesize
    \begin{axis}[
            ymajorgrids=true,
            xtick pos=left,
            ytick pos=left,
            minor y tick num=1,
            minor x tick num=1,
            ytick={21,23,25,27,29,31},
            ymax=31,
            ymin=20,
            xmax=3.5,
            xmin=1.75,
            ylabel=BLEU, xlabel=LAAL (s),
            ylabel shift={-4pt},
            width=4.6cm,
            height=5.2cm,
            compat=newest,
            xtick=data,
            xtick={1.5,2,2.5,3,3.5},
            legend style={at={(0.5,-0.2)},    
                    anchor=north,legend columns=4},   
            legend to name={mylegend},
        ]
        \addplot[color=orange, mark=pentagon*] table[x=AL,y=BLEU]{\NLwaitk};
        \addplot[color=blue, mark=triangle*] table[x=AL,y=BLEU]{\NLcommon};
        \addplot[color=teal, mark=diamond*] table[x=AL,y=BLEU]{\NLedatt};
        \addplot[color=red, mark=*] table[x=AL,y=BLEU]{\NLalignatt};
    \end{axis}
\end{tikzpicture} 
}
\subfigure[en$\rightarrow$pt]{
\begin{tikzpicture}
    \footnotesize
    \begin{axis}[
            ymajorgrids=true,
            xtick pos=left,
            ytick pos=left,
            minor y tick num=1,
            minor x tick num=1,
            ytick={22,24,26,28,30,32},
            ymax=32,
            ymin=22,
            xmin=1.75,
            xmax=3.5,
            ylabel=, xlabel=LAAL (s),
            width=4.6cm,
            height=5.2cm,
            compat=newest,
            xtick=data,
            xtick={1.5,2,2.5,3,3.5},
            legend style={at={(0.5,-0.2)},    
                    anchor=north,legend columns=4},   
            legend to name={mylegend},
        ]
        \addplot[color=orange, mark=pentagon*] table[x=AL,y=BLEU]{\PTwaitk};
        \addplot[color=blue, mark=triangle*] table[x=AL,y=BLEU]{\PTcommon};
        \addplot[color=teal, mark=diamond*] table[x=AL,y=BLEU]{\PTedatt};
        \addplot[color=red, mark=*] table[x=AL,y=BLEU]{\PTalignatt};
        \legend{wait-k, common-prefix, EDAtt, AlignAtt}
    \end{axis}
\end{tikzpicture} 
}
\subfigure[en$\rightarrow$ro]{
\begin{tikzpicture}
    \footnotesize
    \begin{axis}[
            ymajorgrids=true,
            xtick pos=left,
            ytick pos=left,
            minor y tick num=1,
            minor x tick num=1,
            ytick={17,19,21,23,25,27},
            ymax=27,
            ymin=17,
            xmin=1.75,
            xmax=3.5,
            ylabel=, xlabel=LAAL (s),
            width=4.6cm,
            height=5.2cm,
            compat=newest,
            xtick=data,
            xtick={1.5,2,2.5,3,3.5},
            legend style={at={(0.5,-0.2)},    
                    anchor=north,legend columns=4},   
            legend to name={mylegend},
        ]
        \addplot[color=orange, mark=pentagon*] table[x=AL,y=BLEU]{\ROwaitk};
        \addplot[color=blue, mark=triangle*] table[x=AL,y=BLEU]{\ROcommon};
        \addplot[color=teal, mark=diamond*] table[x=AL,y=BLEU]{\ROedatt};
        \addplot[color=red, mark=*] table[x=AL,y=BLEU]{\ROalignatt};
        \legend{wait-k, common-prefix, EDAtt, AlignAtt}
    \end{axis}
\end{tikzpicture} 
}
\subfigure[en$\rightarrow$ru]{
\begin{tikzpicture}
    \footnotesize
    \begin{axis}[
            ymajorgrids=true,
            xtick pos=left,
            ytick pos=left,
            minor y tick num=1,
            minor x tick num=1,
            ytick={12,14,16,18},
            ymax=18,
            ymin=11,
            xmin=1.75,
            xmax=3.5,
            ylabel=, xlabel=LAAL (s),
            width=4.6cm,
            height=5.2cm,
            compat=newest,
            xtick=data,
            xtick={1.5,2,2.5,3,3.5,4,4.5},
            legend style={at={(0.5,-0.2)},    
                    anchor=north,legend columns=4},   
            legend to name={mylegend},
        ]
        \addplot[color=orange, mark=pentagon*] table[x=AL,y=BLEU]{\RUwaitk};
        \addplot[color=blue, mark=triangle*] table[x=AL,y=BLEU]{\RUcommon};
        \addplot[color=teal, mark=diamond*] table[x=AL,y=BLEU]{\RUedatt};
        \addplot[color=red, mark=*] table[x=AL,y=BLEU]{\RUalignatt};
        \legend{wait-k, LA, EDAtt, AlignAtt}
    \end{axis}
\end{tikzpicture} 
}
\ref{mylegend}
\caption{LAAL-BLEU curves for all the 8 language pairs of MuST-C tst-COMMON.\textsc{AlignAtt} is compared to the SimulST policy presented in Section \ref{subsec:comparison}. Latency (LAAL) is computationally aware and expressed in seconds ($s$).}
\label{fig:simul_res}
\end{figure*}
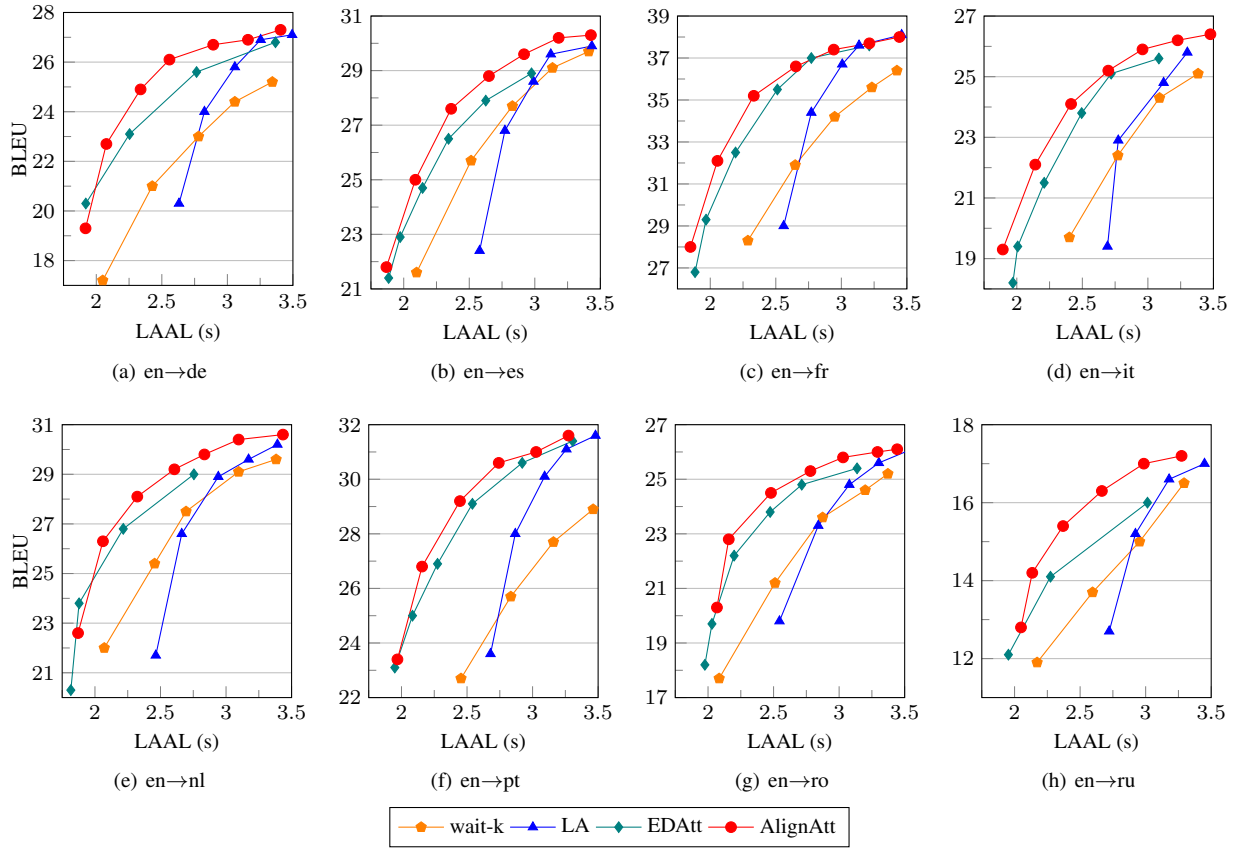

\section{Results}
\label{sec:exps}

In this section, we present the results of our offline systems trained for each language pair of MuST-C v1.0 to show their competitiveness 
%of the model \mn{[QUALE? SICURA CHE SI CAPISCA COSA INTENDI?]} 
compared to the systems published in literature
(Section \ref{subsec:offline_res}) and  the results of the \textsc{AlignAtt} policy compared to the other policies 
presented in Section \ref{subsec:comparison} (Section \ref{subsec:simul_res}).

\subsection{Offline Results}
\label{subsec:offline_res}
To provide an upper bound to the simultaneous performance and show the competitiveness of our models, we present in Table \ref{tab:offline_res} the offline results of the systems trained on all the language pairs of MuST-C v1.0 compared to systems published 
%state-of-the-art systems 
in literature
%reporting all the languages. 
that report results for all languages.
%\mt{Sei sicura che questi soon lo stato dell'arte anche rispetto all'uso di pre-training model? Questo paper e' contemporaneo, ma cita paper che hanno ottime performance https://arxiv.org/pdf/2301.11716.pdf. Se ti vuoi smarcare direi dove non si usano i pre-trained models.} 
As we can see, our offline systems outperform the others on all but 2 language pairs, en$\rightarrow$\{es, fr, it, nl, pt, ro\}, achieving the new state of the 
%art.
art in terms of translation quality.
BLEU gains are more evident for en$\rightarrow$fr and en$\rightarrow$it, for which we obtain improvements of about 1 BLEU point, while they amount to about 0.5 BLEU points for the other languages.

Concerning the other 2 languages (de, ru), our en$\rightarrow$ru model achieves a similar result (18.4 vs 18.5 BLEU) with that obtained by the best model for that language (XSTNet \cite{ye21_interspeech}), with only a 0.1 BLEU drop. 
Moreover, our system reaches a slightly worse but competitive result for en$\rightarrow$de (28.0 vs 28.7 BLEU) compared to STEMM \cite{fang-etal-2022-stemm}, which instead makes use of a
%consistent 
relevant
amount of external speech data, and it also outperforms all the other systems for this language direction. 
%\mt{Ma usa dati addizionali anche per le altre lingue? Se si', come mai solo per il tedesco e' meglio di noi? Usa piu' dati che per le altre lingue?}
On average, our approach 
%results in 
stands out as
the best one even if it does not involve the use of external speech 
%data and 
data\sara{:} %; 
it obtains an average of 29.4 BLEU across languages, which 
%is translated into 
corresponds to
0.5 to 4.6 BLEU improvements compared to the published ST models.

\subsection{Simultaneous Results}
\label{subsec:simul_res}

%Once 
Having
demonstrated the competitiveness of our offline models, we now apply the SimulST policies introduced in Section \ref{subsec:comparison} to the same offline ST model for each language pair of MuST-C v1.0. 
% The results are 
% %depicted
% \mn{shown}
% in Figure \ref{fig:simul_res}.
Figure \ref{fig:simul_res} shows the results in terms of latency-quality trade-off (i.e. LAAL ($\downarrow$) - BLEU ($\uparrow$) curves).

As we can see, our \textsc{AlignAtt} policy is the only policy, together with 
%the \textsc{EDAtt} policy,
\textsc{EDAtt},
capable of reaching a latency lower or equal to 2$s$ 
%among
for all the 8 languages.\footnote{The maximum acceptable latency limit is set between 2$s$ and 3$s$ from most works on simultaneous interpretation \cite{doi:10.1177/002383097501800310,fantinuoli2022defining}.}
%008039ar
Specifically, LA curves start at around 2.5$s$ or more for all the language pairs, even if they are able to achieve high translation quality towards 3.5$s$, with a 1.2 average drop in terms of BLEU 
%among
across languages compared to the offline inference.
Similarly, the wait-k curves start at around 2/2.5$s$ but are not able to reach high translation quality even at high latency (LAAL approaching 3.5$s$), therefore scoring the worst results. 
Compared to these two policies, \textsc{AlignAtt} shows a LAAL reduction of up to 0.8$s$ 
%for
compared to
LA and 0.5$s$ 
%for
compared to
wait-k.
Despite achieving lower latency as \textsc{AlignAtt}, the \textsc{EDAtt} policy achieves worse
translation quality at almost every latency regime compared to our policy, with drops of up to 2 BLEU
%among
points across
languages. These performance drops are particularly evident for en$\rightarrow$de and en$\rightarrow$ru, where the latter represents the most difficult language pair also in offline ST (it is the only language with less than 20 BLEU on Table \ref{tab:offline_res}). 
%This indicates that our policy can better handle Russian, the only Slavic target language present in MuST-C v1.0, suggesting its effective applicability to other alphabets.
The evident differences in the \textsc{AlignAtt} and \textsc{EDAtt} policy behaviors, especially in terms of translation quality, prove that, despite both exploiting attention scores as a source of information, the decisions taken by the two policies are intrinsically different. Moreover, \textsc{AlignAtt} is the closest policy to achieving the offline results of Table \ref{tab:offline_res}, with less than 1.0 BLEU average drop versus 1.8 of
%the 
\textsc{EDAtt}.
%policy 
%compared to the results of Table . 
%\mt{Ha senso mettere le performance dell'offline model nei plots? Non vorrei che pero' schiacciassero troppo le curve.}

% Overall, we can conclude that the \textsc{AlignAtt} policy achieves a reduced latency compared to both wait-k and LA and an improved translation quality compared to \textsc{EDAtt} on all the 8 languages of MuST-C v1.0, therefore representing the new state-of-the-art SimulST policy applicable to offline ST models.
%In light of the above, we 
We can 
conclude that, on all the 8 languages of MuST-C v1.0, the \textsc{AlignAtt} policy achieves a lower latency compared to both wait-k and LA, and an improved translation quality compared to \textsc{EDAtt}, therefore representing the new state-of-the-art SimulST policy applicable to offline ST models.

\section{Conclusions}
%In this paper, we 
We presented \textsc{AlignAtt}, a novel policy for SimulST that leverages the audio-translation alignments obtained from the cross-attention scores to guide an offline-trained ST model during simultaneous inference. Results on all 8 languages of MuST-C v1.0 showed the effectiveness of our policy compared to the existing ones, with gains of 2 BLEU and a latency reduction of 0.5-0.8$s$, achieving the new state of the art. 
Code, offline ST models, and simultaneous outputs are released open source to help the reproducibility of our work. 
%\sara{Code, offline ST models, and simultaneous outputs are released open source.
%Future works can be devoted to exploring more languages and domains of application of the proposed policy.}

\bibliographystyle{IEEEtran}
\bibliography{mybib}

\end{document}